\documentclass{article} 
\usepackage{iclr2021_conference,times}

\usepackage{amsmath,amsfonts,bm}

\def\eqref#1{equation~\ref{#1}}

\def\1{\bm{1}}

\DeclareMathAlphabet{\mathsfit}{\encodingdefault}{\sfdefault}{m}{sl}
\SetMathAlphabet{\mathsfit}{bold}{\encodingdefault}{\sfdefault}{bx}{n}

\usepackage{hyperref}
\usepackage{url}
\usepackage{graphicx}

\iclrfinalcopy

\title{Variational Autoencoder for Anti-Cancer\\ Drug Response Prediction}
\author{
Hongyuan Dong \footnotemark[1] \\
Harbin Institute of Technology at Weihai\\
\texttt{170400305@stu.hit.edu.cn} 
\And
Jiaqing Xie \thanks{Equal Contribution, source code location: https://github.com/JIAQING-XIE/Machine-Learning-in-Genomes/tree/main} \\
University of Edinburgh\\
\texttt{s2001696@ed.ac.uk} 
\And
Zhi Jing \\
Sun Yat-sen University\\
\texttt{jingzh5@mail2.sysu.edu.cn} 
\And
Dexin Ren \\
University of Arizona\\
\texttt{dexinren@email.arizona.edu} 
}

\begin{document}

\maketitle

\begin{abstract}
Cancer is a primary cause of human death, but discovering drugs and tailoring cancer therapies are expensive and time-consuming. We seek to facilitate the discovery of new drugs and treatment strategies for cancer using variational autoencoders (VAEs) and multi-layer perceptrons (MLPs) to predict anti-cancer drug responses. Our model takes as input gene expression data of cancer cell lines and anti-cancer drug molecular data and encodes these data with our {\sc {GeneVae}} model, which is an ordinary VAE model, and a rectified junction tree variational autoencoder ({\sc JTVae}) model, respectively. A multi-layer perceptron processes these encoded features to produce a final prediction. Our tests show our system attains a high average coefficient of determination ($R^{2} = 0.830$) in predicting drug responses for breast cancer cell lines and an average $R^{2} = 0.845$ for pan-cancer cell lines. Additionally, we show that our model can generate effective drug compounds not previously used for specific cancer cell lines.
\end{abstract}

\section{Introduction and Related Works}

The discovery of new drugs and the customization of cancer therapy remain difficult problems. Cancer drugs are a widely used primary treatment. However, development of these drugs is expensive and time-consuming, and it is difficult to tailor therapy to individual patients. We propose a generative model for accurate prediction of anti-cancer drug response to help with this critical need.

The effectiveness of cancer drugs is highly dependent on the genomic and transcriptomic profile of the specific cancers.\cite{yang2012genomics} Some researchers have predicted drug response using gene expression data. \cite{chiu2019predicting} built deep neural networks to combine gene expression with mutation profiles to make predictions, and \cite{geeleher2014clinical} implement a ridge regression model on before-treatment gene expression data to predict response of chemotherapy. Our strategy incorporates both gene expression and anti-cancer drug molecular data to predict responses of different drugs on various cancer cell lines. 

Auto-encoders have been used widely to extract low dimensional features from unlabeled data, but they are not very robust, with slight variances in the encoded vector sometimes leading to huge differences in the reconstructed data. Other feature extraction methods such as convolutional neural networks (CNNs)\cite{chang2018cancer} and graph convolutional networks (GCNs)\cite{liu2020deepcdr}, can be used to extract features from unlabeled gene genomic data and drug molecular data. But these methods can not function as the generative models of great significance in drug discovery. Therefore, we employ a variational autoencoder ({\sc VAE}),\cite{kingma2013auto} which models the distribution of latent features instead of producing specific latent features directly, to extract features from unlabeled gene expression data and drug molecular data. An ordinary VAE model (GeneVAE) with its encoder and decoder both composed of 2-layer neural networks is implemented for gene expression profile of cancer cell lines. For analyzing anti-cancer drugs, we adopt a junction tree VAE (JTVAE)\cite{jin2018junction} model to transform the molecular graphs into valid substructures to extract their low dimensional features. Using the encoded low-dimensional features of the gene expression and drug molecular data, we implement a multi-layer perceptron (MLP) to combine the extracted features and produce the final result, which is the $ln(IC_{50})$ value of the target anti-cancer drug used against a specific cancer cell line. Differing from previous works with models restricted to specific drugs,\cite{10.1093/neuonc/not216, yuasa2011biomarkers, imamura2013urine} our model can take any organic compound as input to predict its usefulness in treatment. 

JTVAE is also a generative model and outperforms many previous approaches\cite{kusner2017grammar,li2018learning,simonovsky2018graphvae} in reconstructing molecules. Drug compounds generated by JTVAE are always valid, making it extremely powerful for discovering new anti-cancer drugs. Our research shows that encoded features of drugs can be randomly sampled, with the well-performing features decoded by JTVAE to reveal a large number of valid compounds effective for cancer therapy. Differing from previous works such as \cite{mendez2020novo}, which only focuses on generating effective drugs with Generative Adversarial Networks (GAN) and VAE, our model can generate effective drug compounds as well as predict drug response on different cancer cell lines, offering promise in reducing the development cost of new drugs. 
\vspace{-1em}
\section{Methods}
\vspace{-1em}
\subsection{Datasets}

We use gene expression data of 1021 cancer lines with 57820 genes provided by the CCLE. \cite{barretina2012cancer} Each cell line belongs to a specific cancer type. Specifically, we choose breast cancer as our primary research object, and later test our model on pan cancer cell lines. We prepare the ZINC dataset for molecular structure data of organic compounds to train the JTVAE model. Molecular structure data is given in simplified molecular-input line entry system (SMILES) strings.  From the ZINC data set, we select 10000 SMILES strings to train our JTVAE model, due to the limitation of GPU resources. Training all the samples requires an over 8GB GPU memory size to hold model parameters so we down-sampling them.
We use drug response data from the Genomics of Drug Sensitivity in Cancer (GDSC) project,\cite{yang2012genomics} which contains response data for cancer drugs against numerous cancer cell lines. We obtain molecular data for the drugs from the PubChem dataset with their unique PubChem ID available from the GDSC dataset. In total, we have 3358 pieces of drug response data for breast cancer cell lines where gene expression data and drug molecular structure are available. We also use the CGC dataset,\cite{chang2018cancer} which classifies different genes into two tiers. One tier is for the genes that are closely associated with cancers and have a high probability to mutate into cancers that change the activity of the gene product. The other tier includes genes that possibly play a strong role in cancer but lack evidence. Genes in both tiers are highly relevant with cancer, which is why we incorporate both tiers.

\vspace{-1em}
\subsection{Gene expression VAE (geneVAE)}
GeneVAE(\ref{appendix:geneVAE}) extracts latent vectors from CCLE gene expression data, with the extracted latent vectors used for drug response prediction. GeneVAE is an ordinary VAE based on fully connected neural networks. For the encoder, we use 2-layer fully connected neural networks for forward propagation with a batch-norm layer before activation:
Latent variables $\textup{z}_{g} \sim \mathcal{N}\left(\mu_{g},\sigma_{g}^{2}\right)$.
$\mu_{g}$ is the computed mean value of this Gaussian distribution. Similarly, $\sigma_{g}$ is computed by another 2-layer neural network with the same architecture as $\mu_{g}$. The latent vector $z_{g}$ is randomly sampled from $\mathcal N (\mu_{g},\sigma_{g})$.
The decoder architecture is also a 2-layer fully connected neural network. The sizes of both encoder layers are set as 256, while the sizes of both decoder layers are set to match input data. When encoding gene expression data into latent vectors, we take $\mu_{g}$ as encoded features instead of sampling these vectors from a Gaussian distribution.

\vspace{-1em}
\subsection{JTVAE}
JTVAE(\ref{appendix:JTVAE}) \cite{jin2018junction} consists of a graph VAE and a tree VAE. Molecules are decomposed as junction trees where nodes are valid molecular substructures. The decomposed junction tree is encoded with a tree VAE while the original molecular graph is encoded with a graph VAE. When generating molecules, the decoder of the tree VAE reconstructs the junction tree of the molecule, and the decoder of the graph VAE provides complementary connectivity information to reproduce the full molecular graph.

\vspace{-1em}
\section{Experiments}
\vspace{-1em}
\subsection{Results on breast cancer}
We prepared several models and tested them on breast cancer cell lines, with the results showing that the VAE and CGC datasets contributed to more accurate predictions. We selected 2 metrics to evaluate performance. The coefficient of determination ($R^{2}$ score) and RMSE evaluated the discrepancy between our predicted drug response and true drug response. We prepared 6 models, and each model was trained and tested on randomly divided training and test datasets for 15 times. Their average performance on test dataset is shown in Table~\ref{table 1}. Among these models, the first 5 models targeted breast cancer, and the last one is tested on pan cancer cell lines. The models used were:
$\mathbf{1)~CGC+SVR:}$ An SVR model trained on drug molecular structure data encoded by JTVAE and gene expression data filtered by the CGC dataset.
$\mathbf{2)~CGC+VAE+SVR:}$ An SVR model trained on drug molecular data encoded by JTVAE, along with gene expression data filtered by the CGC dataset and encoded by geneVAE.
$\mathbf{3)~CGC+MLP:}$ An MLP model trained on drug molecular structure data encoded by JTVAE and gene expression data filtered by the CGC dataset.
$\mathbf{4)~RAW+VAE+MLP:}$ An MLP model trained on drug molecular data encoded by JTVAE and raw gene expression data (not filtered by CGC dataset) encoded by geneVAE.
$\mathbf{5)~CGC+VAE+MLP:}$ An MLP model trained on drug molecular structure data encoded by JTVAE along with gene expression data filtered by CGC and encoded by geneVAE.
$\mathbf{6)~CGC+VAE+MLP:}$ An MLP model trained on drug molecular structure data encoded by JTVAE along with gene expression data filtered by CGC and encoded by geneVAE. This model was trained on pan cancer dataset.

\begin{figure}[!htpb]
\label{results}
\centering
\begin{minipage}[t]{0.32\textwidth}
\centering
\includegraphics[width=4.4cm]{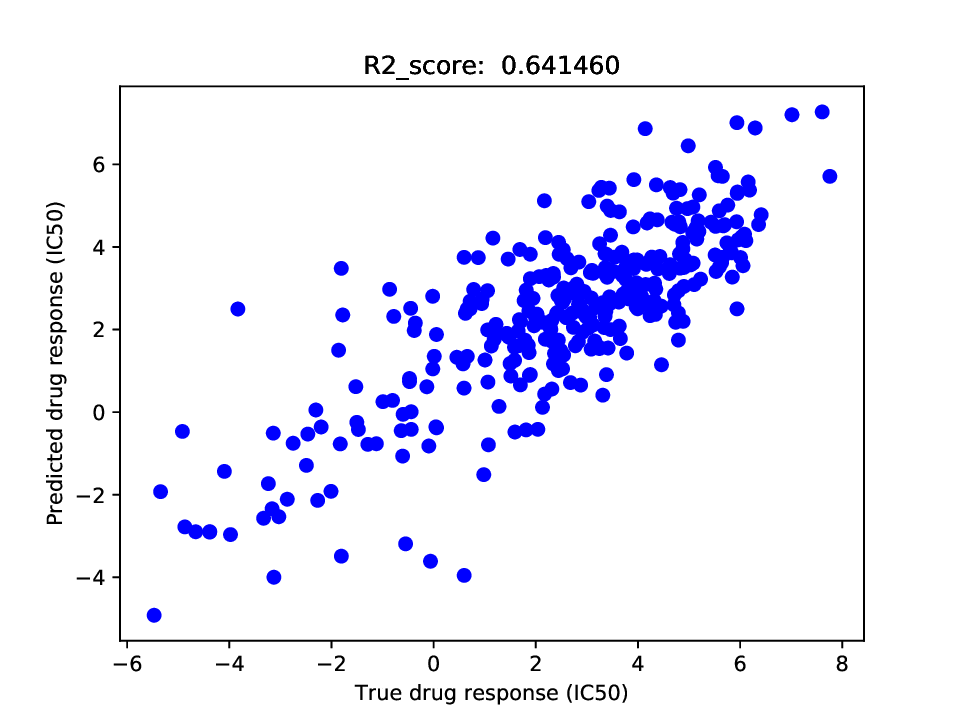}
\caption{CGC+SVR}
\label{cgc+svr}
\end{minipage}
\begin{minipage}[t]{0.32\textwidth}
\centering
\includegraphics[width=4.4cm]{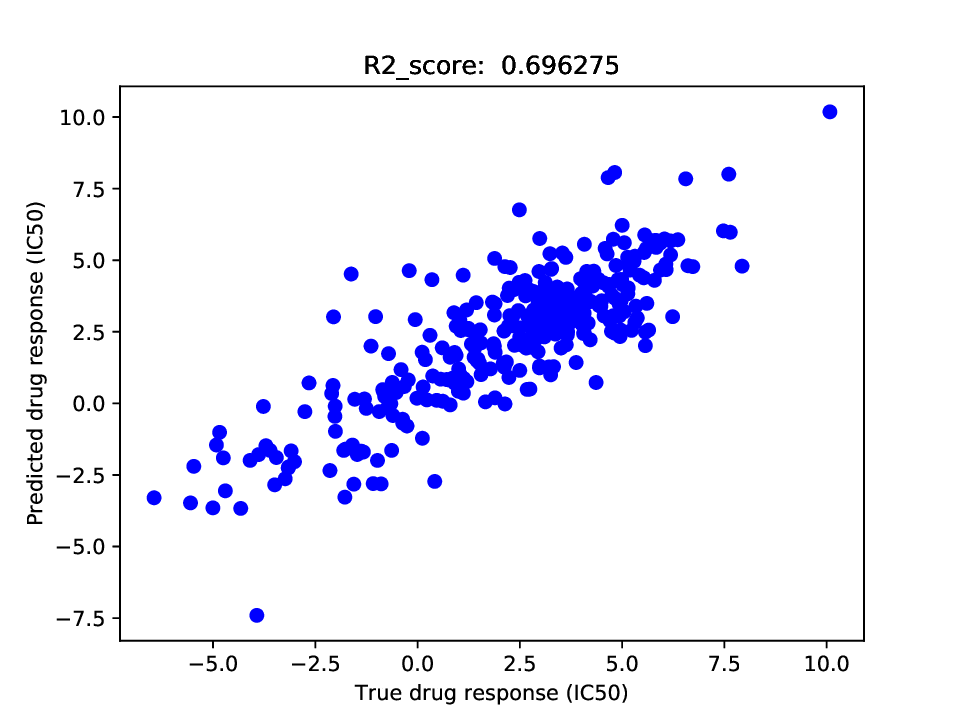}
\caption{CGC+VAE+SVR}
\label{cgc+vae+svr}
\end{minipage}
\begin{minipage}[t]{0.32\textwidth}
\centering
\includegraphics[width=4.4cm]{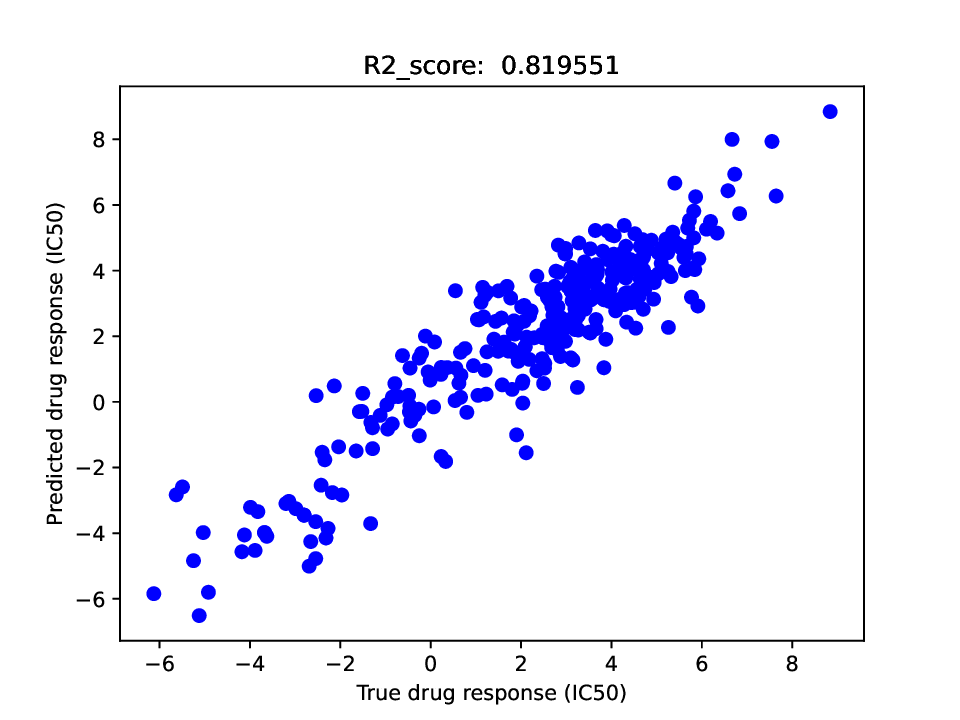}
\caption{CGC+MLP}
\label{cgc+mlp}
\end{minipage}
\begin{minipage}[t]{0.32\textwidth}
\centering
\includegraphics[width=4.4cm]{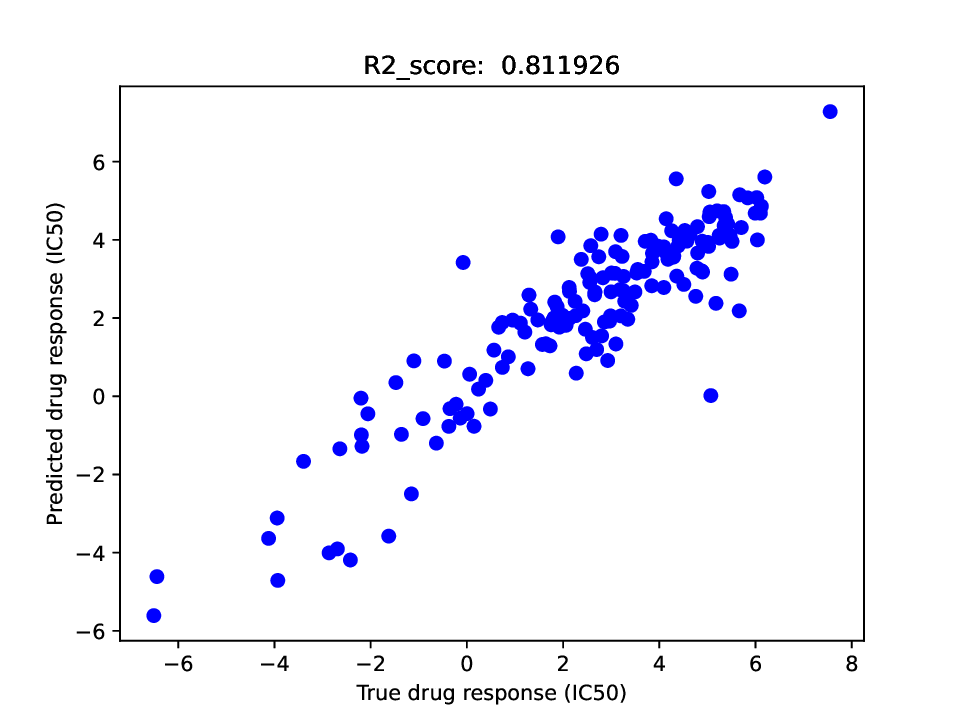}
\caption{Raw+VAE+MLP}
\label{raw+vae+mlp}
\end{minipage}
\begin{minipage}[t]{0.32\textwidth}
\centering
\includegraphics[width=4.4cm]{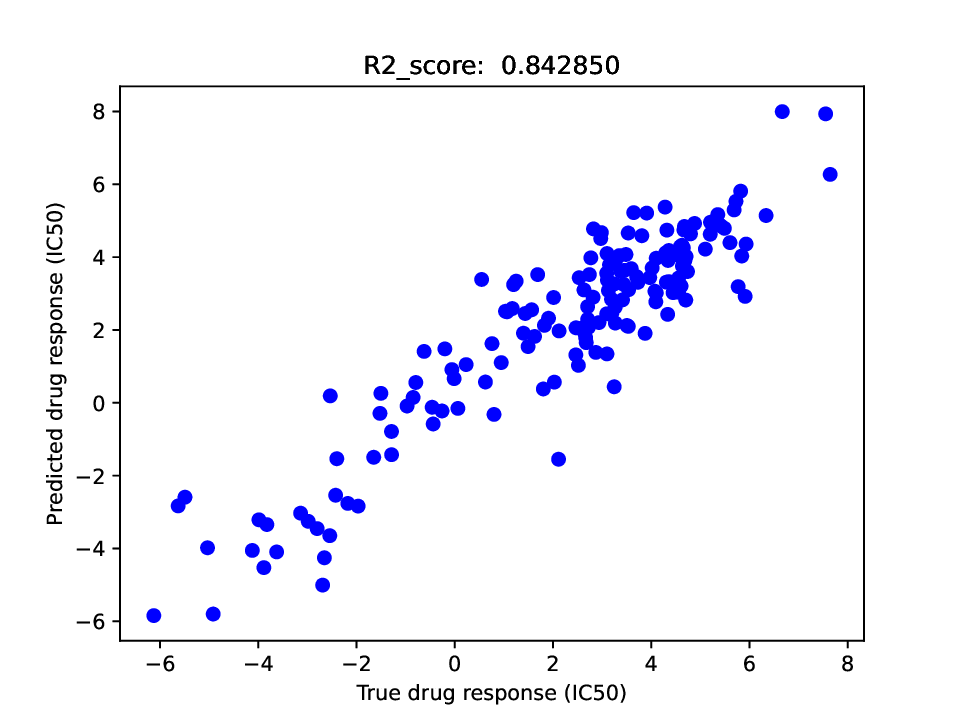}
\caption{CGC+VAE+MLP}
\label{cgc+vae+mlp}
\end{minipage}
\begin{minipage}[t]{0.32\textwidth}
\centering
\includegraphics[width=4.4cm]{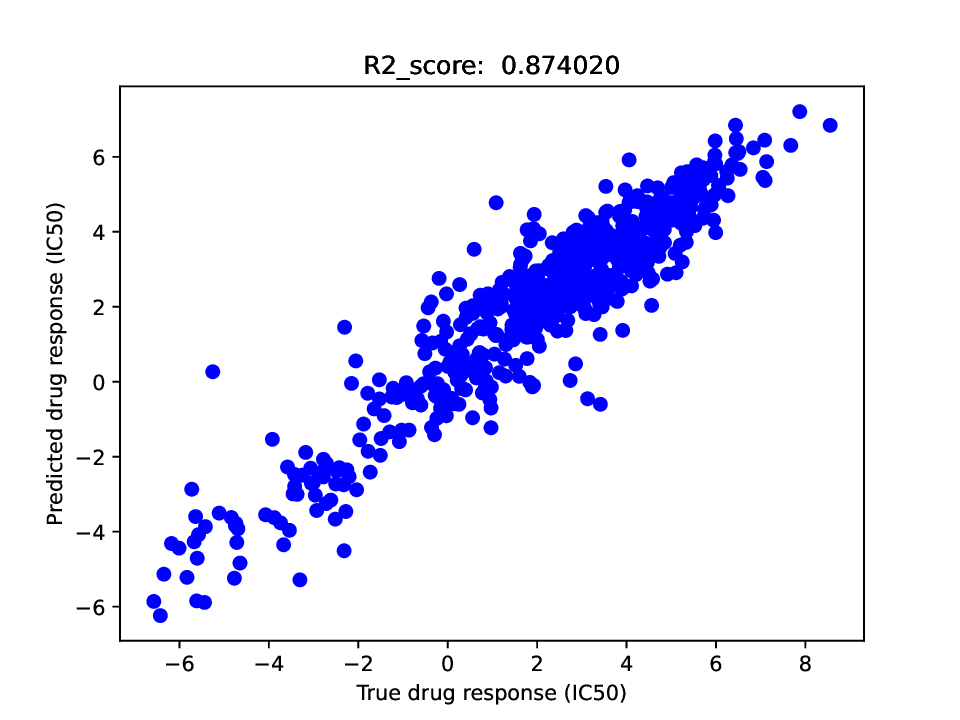}
\caption{CGC+VAE+MLP (tested on pan cancer cell lines)}
\label{cgc+vae+mlp (pan)}
\end{minipage}
\end{figure}

Table \ref{table 1} presents the performance comparison between our proposed models. Scatter plots illustrating the relationship between true and predicted values from the models on the test sets are shown in Figures~\ref{cgc+svr} to \ref{cgc+vae+mlp (pan)}. Results indicate MLP and VAE improved the performance of our models significantly. The $\mathbf{ CGC+MLP}$ model outperformed the $\mathbf{ CGC+SVR}$ model by 0.164 on the $R^{2}$ score, and the $\mathbf{ CGC+VAE+MLP}$ model performed even better than the $\mathbf{ CGC+MLP}$ model with a 0.008 higher $R^{2}$ score. Filtering out an important gene subset with the CGC dataset was also essential to the performance of our models. For example, the $\mathbf{ CGC+VAE+MLP}$ model on breast cancer cell lines reached an $R^{2}$ score of 0.830, outperforming the $\mathbf{ RAW+VAE+MLP}$ model by 0.025. The 
gap between $\mathbf{ CGC+MLP}$ and $\mathbf{ CGC+VAE+MLP}$ is small because of similar model structures. However, the difference is also evident since VAE is more likely to present a true latent space of the drug representation(explained by a small KL divergence of VAE), which will lead to a 
correct generation or reconstruction of drugs while MLP does not have this function.
\begin{table}[!htp]
\caption{Performance of the 6 proposed models on breast and pan cancer datasets}
\begin{center}
\label{table 1}
\begin{tabular}{llll}
\hline
Models & Cancer type & $R^{2}_{test}$ & $RMSE_{test}$\\
\hline
CGC + SVR & Breast & 0.658 & 1.582\\
CGC + VAE + SVR & Breast & 0.692 & 1.491\\
CGC + MLP & Breast & 0.822 & 1.133\\
RAW + VAE + MLP & Breast & 0.805 & 1.163\\
CGC + VAE + MLP & Breast & $\mathbf{0.830}$ & 1.130\\
CGC + VAE + MLP & Pan cancer & $\mathbf{0.845}$ & 1.080\\
\hline
\end{tabular}
\end{center}
\end{table}

\subsection{Test on pan cancer}

We further tested our model on the pan-cancer cell lines from the CCLE dataset. The total number of cell lines was 1021, and we used 13605 pieces of drug response data to train and test our model. The $\mathbf{ CGC+VAE+MLP}$ model, the best performer with breast cancer cell lines, achieved an even higher $R^{2}$ score of 0.845 on pan-cancer cell lines. The reason might be resulted from the complicated mechanism of breast cancer when compared to other cancer types, which is needed to be verified further.
\vspace{-1em}
\subsection{Effective drug compound generation}
\begin{figure*}[!htpb]
\setlength\abovedisplayskip{2pt}
\setlength\belowdisplayskip{2pt}
\centering
\includegraphics[scale=0.26]{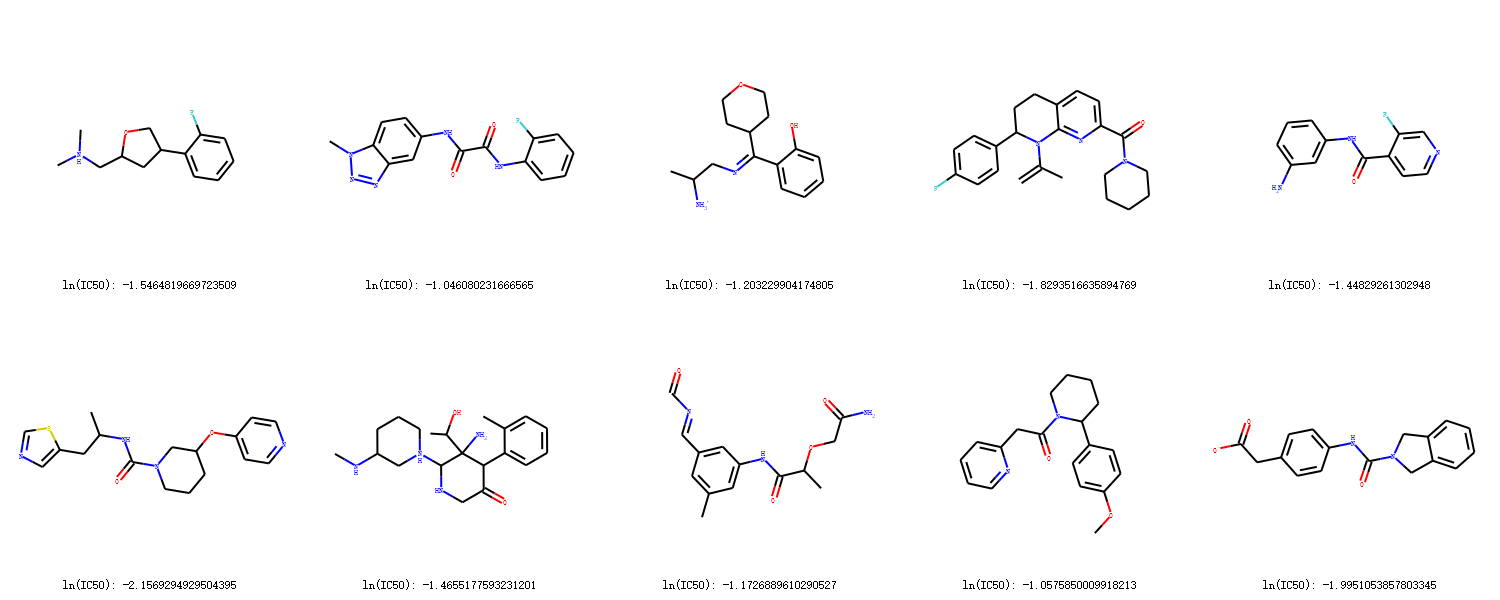}
\caption{Ten effective drugs whose $ln(IC_{50})$ values on cancer cell line HCC1187 are below $-1.0$}
\label{Effective drugs}
\end{figure*}

Compared with other representation learning methods on molecules, JTVAE has the advantage of reconstructing some valid drugs, making it powerful in generating drugs for specific cancer cell lines. In our experiments, we used the breast cancer cell line HCC1187 as an example to demonstrate how our model generated customized and effective drug compounds for a given cancer cell line. First, we sampled several 56-dimension vectors to match the dimensionality of the latent vectors encoded by JTVAE from the Gaussian distribution $\mathcal N(\mu,\sigma^{2})$, where $\mu=0$ and $\sigma=7$. The randomly sampled drug vectors were concatenated with the encoded latent vector of gene expression profile of HCC1187. The MLP model ingested the concatenated vectors and produced a prediction. We set the threshold of effective drugs as $-1.0$ $ln(IC_{50})$ value. If the $ln(IC_{50})$ value of a randomly generated drug latent vector was below $-1.0$, it was considered to be effective on HCC1187. Also, the threshold could be set as $-1.5$, $-2.0$ etc. to produce more effective generated drugs. We selected 10 generated drug latent vectors whose $ln(IC_{50})$ values on HCC1187 were below $-1.0$ and decoded them with the JTVAE model. Results are shown in Figure~\ref{Effective drugs}. As JTVAE always decode drug latent vectors into valid compounds, these decoded drug compounds, which might have not been used as cancer drugs previously, showed promise for cancer treatment.Some further experiments should be validated on the effectiveness of generated drugs.
\vspace{-1em}
\section{Conclusions}
Since it is extremely expensive and time-consuming to develop new cancer drugs and propose personalized cancer treatment therapies, we seek to use VAE and MLP models to produce accurate predictions of drug efficacy and to generate effective drugs for given cancer cell lines. We use the JTVAE\cite{jin2018junction}) model and construct the geneVAE model to process SMILES drug data and gene expression profiles of cancer cell lines, respectively. JTVAE and geneVAE encode these data into representative low dimensional features, which the MLP model uses to make drug efficacy predictions. Our comparison of models using both breast cancer and multiple cancer cell lines show that encoding data using VAE and curating a gene subset with the CGC dataset contribute to better performance. Our best $\mathbf{CGC+VAE+MLP}$ model achieves an encouraging coefficient of determination value (0.845 $R^{2}$ score on pan cancer and 0.830 on breast cancer). In addition, we demonstrate that our model works as a generative model to generate effective cancer drugs for given cancer cell lines. We have also explored the latent vectors encoded by geneVAE and JTVAE to demonstrate the validity of our pipeline.

\bibliography{iclr2021_conference}
\bibliographystyle{iclr2021_conference}

\appendix
\section{Related work}
\subsection{Feature dimensionality reduction}
Encoding features into lower dimensions is commonly used for representation learning tasks. The reduction in feature dimensionality removes a large amount of redundant information to facilitate the analysis. Supervised learning methods can select features which are most relevant with the task. As examples, \citeauthor{wenric2018using}\cite{wenric2018using} use random forests to determine gene importance in RNA sequence case-control studies, and \citeauthor{liu2018feature}\cite{liu2018feature} implement support vector machines (SVM) with double RBF-kernels to filter out irrelevant gene features. Unsupervised learning methods, such as PCA and hierarchical learning are useful in explaining the group features of genes while reducing the feature dimensionality.\cite{huang2006unsupervised} Auto-encoders based on neural networks also learn to encode original data into low-dimensional features without supervision and have been widely used to extract low-dimensional features. Though auto-encoders outperform traditional methods, their robustness is unpredictable; slight variances in the encoded vector can lead to huge differences in the reconstructed data. Variational auto-encoders,\cite{kingma2013auto} which add noise to the encoded features to build a more robust auto-encoder model, have been proposed to overcome this weakness.

\subsection{Variational auto-encoders with gene profiles}
Much work has been done on encoding gene expression data into representative low-dimensional features. Neural networks, such as MLPs and convolutional neural networks (CNNs), can encode gene features effectively. \citeauthor{chang2018cancer}\cite{chang2018cancer} use a CNN to encode gene mutation and drug molecular data, and \citeauthor{oskooei2018paccmann}\cite{oskooei2018paccmann} implement attention-based neural networks to produce explainable encoded features. An encoder-decoder structure\cite{chiu2019predicting} extends ordinary MLPs that are also able to reconstruct the original input. The bottleneck layer represents the latent features encoded by autoencoders. Recently, the VAE,\cite{kingma2013auto} which modifies ordinary auto-encoders to improve robustness, has been used frequently in pre-trained models for gene expression data. \citeauthor{gronbech2018scvae}\cite{gronbech2018scvae} use a VAE to estimate expected gene expression level, and \citeauthor{rampasek2017dr}\cite{gronbech2018scvae} implement VAE models to analyze pre- and post-treatment gene expression profiles to make anti-cancer drug response prediction. We also incorporate the VAE model to process gene expression data. Latent features of gene expression data provide representative information about cancer cell lines that enables our model to predict drug responses for different cancer tissues.

\subsection{Representation learning on graphs for drug molecular features}
Drug molecular features can be represented as graphs and processed by deep neural networks. \citeauthor{duvenaud2015convolutional}\cite{duvenaud2015convolutional} build graph CNNs on circular fingerprints of molecules. \citeauthor{liu2020deepcdr}\cite{liu2020deepcdr} implement uniform graph convolutional neural networks (UGCNs) to extract representative features from drug molecular data. \citeauthor{gilmer2017neural}\cite{gilmer2017neural} use a message passing neural network (MPNN) for molecular property prediction.

In addition, attention mechanisms can be used with RNN and CNN models\cite{manica2019toward,oskooei2018paccmann} to encode drug molecular data, learning attention weights by multihead-attention or self-attention to produce explainable encoded features. The VAE is also widely used in tasks that require the models to be generative. \citeauthor{kusner2017grammar}\cite{kusner2017grammar} propose a grammar-based VAE and use parse trees to produce more valid generated output, and \citeauthor{simonovsky2018graphvae}\cite{simonovsky2018graphvae} label the nodes and bonds in molecules to form a graph structure and apply the VAE model on it. \citeauthor{li2018learning}\cite{li2018learning} use a graph-structured VAE model to generate molecules matching the statistics of the original dataset.

In order to avoid generating atoms one by one, which often leads to invalid output in drug design, \citeauthor{jin2018junction}\cite{jin2018junction} propose a JTVAE that decomposes molecules into valid substructures and generates compounds from a vocabulary of valid components. As a result, the molecules generated by JTVAE are always valid. For this reason, we make use of the JTVAE as our pre-trained model to encode molecular drug data and generate effective drugs for cancer cell lines.

\subsection{Drug response prediction methods}
Drug response prediction is a supervised regression task. Support vector regression (SVR) and random forest regressors are basic algorithms to perform regression. Recently, deep neural network methods have become popular in drug efficacy prediction. \citeauthor{chiu2019predicting}\cite{chiu2019predicting} build deep neural networks to analyze gene expression and mutation profiles to make predictions, and \citeauthor{chang2018cancer}\cite{chang2018cancer} use CNN-based methods on gene mutation profiles and drug molecular data. \citeauthor{liu2020deepcdr}\cite{liu2020deepcdr} also use gene mutation data and drug molecular data and apply CNNs and UGCNs to make predictions, while \citeauthor{oskooei2018paccmann}\cite{oskooei2018paccmann} implement attention-based neural networks for gene expression and molecular drug data to make explainable predictions. In our approach, we implement a MLP model for encoded gene expression and drug molecular data to make predictions.

\section{Materials and methods}
In this section we present our strategy for processing the datasets along with our model implementation. Our model takes as input the gene expression data of a cancer cell line and the SMILES representation of an anti-cancer drug, and produce a drug response prediction in terms of ${ln}(IC_{50})$. The model consists of geneVAE, an ordinary VAE, to extract features from the gene expression data, a JTVAE to extract features from the molecular drug data, and an MLP model to produce a final prediction.

\subsection{Data}
\subsubsection{Gene expression data}
We use gene expression data of 1021 cancer lines with 57820 genes provided by the CCLE.\cite{barretina2012cancer} Each cell line belongs to a specific cancer type. Specifically, we choose breast cancer as our primary research object, and later test our model on pan cancer cell lines. After filtering by the key word token {\ttfamily[BREAST]}, we select 51 breast cancer cell lines from this dataset: {\ttfamily[AU565\_BREAST],\ttfamily[BT20\_BREAST], \ttfamily[ZR7530\_BREAST]}, and so on. Gene expression data is given by $G \in R^{g \times c}$, where $g$ is the number of genes and $c$ is the number of cancer cell lines. The elements of matrix $G$ are ${log}_{2}(t_{pm} + 1)$, where $t_{pm}$ is the transcriptome per million (tpm) value of the gene in the corresponding cell line.

 We select 51 breast cancer cell lines from the CCLE data set and remove expression data of genes which are not in the CGC dataset. Each gene expression entrance with a mean of $\mu $ which is less than 1 or standard deviation $ \sigma $ which is less than 0.5 is also removed due to their low relevance to cancer cell lines.\cite{chiu2019predicting} Our final set contains gene expression data of 597 genes in 51 breast cancer cell lines.

\subsubsection{Anti-cancer drug molecular structure data}
In our research, we prepare the ZINC dataset for molecular structure data of organic compounds to train the JTVAE model. Molecular structure data is given in simplified molecular-input line entry system (SMILES) strings. The SMILES representation is often used to define drug structures\cite{jin2018junction,kusner2017grammar,simonovsky2018graphvae,chang2018cancer,manica2019toward,oskooei2018paccmann,liu2018constrained,tsubaki2019compound} and are widely used as inputs for drug structure prediction. The SMILES representation simplifies obtaining the embeddings from the vocabulary parsing library we have generated. From the ZINC data set, we select 10000 SMILES strings to train our JTVAE model. The number of SMILES strings used for pre-training is far larger than the actual number of 222 drugs in the processed GDSC dataset. The reason is that we would like to improve our model's robustness with all drugs, not just anti-cancer drugs.

\subsubsection{Drug response data}
We use drug response data from the Genomics of Drug Sensitivity in Cancer (GDSC) project,\cite{yang2012genomics} which contains response data for cancer drugs against numerous cancer cell lines. Data from the GDSC data set is given by a matrix $IC_{CCLE} \in R^{d \times c}$, where $d$ is number of drugs, and $c$ is the number of cancer lines. The elements in this matrix are ${ln}(IC_{50})$ values, where $IC_{50}$ is the half maximal inhibitory concentration value of the drugs used against specific cancer cell lines. We obtain molecular data for the drugs from the PubChem dataset with their unique PubChem ID available from the GDSC dataset. In total, we have 3358 pieces of drug response data for breast cancer cell lines where gene expression data and drug molecular structure are available.

\subsection{Variational Auto-encoder}

The variational auto-encoder is a generative model, modeling the complicated conditional distribution of latent features with an inference network (encoder) and a generative network (decoder). In describing the VAE, $q(z|x; \phi)$ is the distribution of approximated latent variables with parameter set $\phi$, where $p(x|z; \theta)$ is the conditional probability distribution computed by the generative network (decoder) with parameter set $\theta$. The aim of VAE is to find the parameters $\phi^{*}$ and $\theta^{*}$ to maximize $\textup{ELBO}(\phi, \theta)$\cite{kingma2013auto}:
\begin{equation}
\begin{split}
\setlength\abovedisplayskip{5pt}
\setlength\belowdisplayskip{5pt}
\phi^{*}, \theta^{*} =\arg\underset{\phi, \theta}{\max}\, \mathbb{E}_{q(z|x; \phi)}[\log p(x|z; \theta)] - \textup{KL}(q(z|x; \phi)||p(z; \theta))
= \arg\underset{\phi, \theta}{\max}, \textup{ELBO}(\phi, \theta).\\
\end{split}
\label{eq1}
\end{equation}
In a VAE, the prior distribution of latent variables $p(z; \theta)$ is approximated as a normal Gaussian distribution, and the posterior $q(z|x; \phi)$ is also expected to follow a Gaussian distribution. The conditional probability distribution $p(x|z; \theta)$ should follow a multivariate Gaussian distribution. Given these suppositions, the estimator for this model and datapoint $\textup{x}$ is\cite{kingma2013auto}):
\begin{equation}
\begin{split}
\setlength\abovedisplayskip{5pt}
\setlength\belowdisplayskip{5pt}
\textup{ELBO}(\phi, \theta) = \frac{1}{2}\sum_{j = 1}^{J}\left(1 + \textup{log}((\sigma_{j})^{2}) - (\mu_{j})^{2} - (\sigma_{j})^{2}\right) +
\frac{1}{L} \sum_{j = 1}^{L}\textup{log}p(\textup{x}|z^{(j)};{\theta}),
\end{split}
\label{eq2}
\end{equation}
where $J$ is the dimensionality of latent variable $z$, and $L$ is number of samples used to compute $\mathbb{E}_{q(z|x; \phi)}[\log p(x|z; \theta)]$ approximately.
In practice, the total loss of the VAE model is set to be the opposite number of ELBO, which satisfies the gradient descent requirement. Because $\textup{z} \sim \mathcal{N}\left(\mu,\sigma^{2}\right)$, one valid reparameterization of $z$ to enable back propagation is $\textup{z} = \mu + \epsilon\sigma$, where $\epsilon \sim \mathcal N\left(0,1\right)$.

\subsection{Gene expression VAE (geneVAE)}
\label{appendix:geneVAE}
\begin{figure}[!tpb]
\centering
\includegraphics[width=8cm]{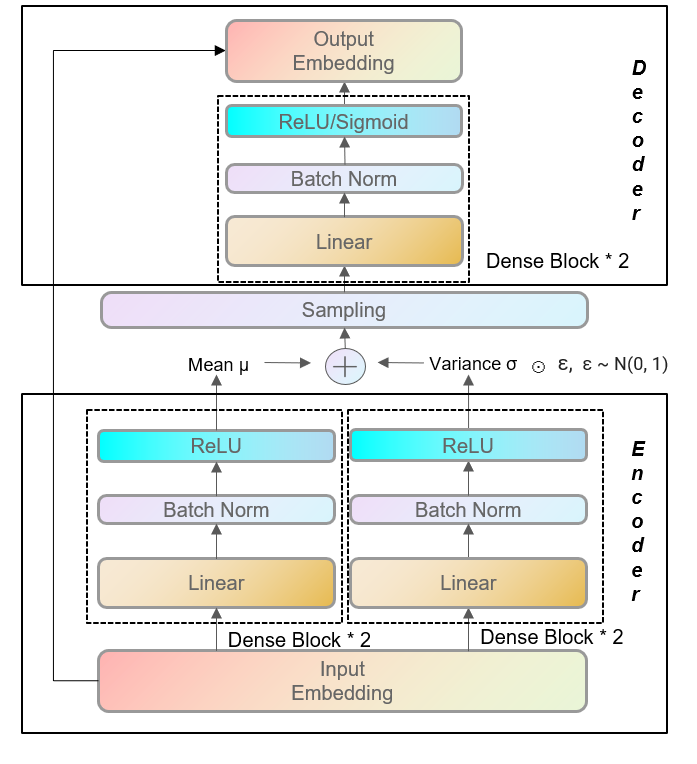}
\caption{The architecture of geneVAE. The encoder computes parameters $\mu$ and $\sigma$ of the Gaussian distribution $\mathcal N\left(0,1\right)$ with separate dense blocks. Sampled latent vectors are processed by the decoder where the first layer uses ReLU activation and the second layer uses sigmoid activation to reconstruct the input}
\label{fig GeneVAE}
\end{figure}

GeneVAE extracts latent vectors from CCLE gene expression data, with the extracted latent vectors used for drug response prediction. geneVAE is an ordinary VAE based on fully connected neural networks. For the encoder, we use 2-layer fully connected neural networks for forward propagation with a batch-norm layer before activation:
\begin{equation}
\setlength\abovedisplayskip{2pt}
\setlength\belowdisplayskip{2pt}
h_{1} = \tau\left(\textup{BN}\left(\mathbf{W}_{1}^{T}G + \mathbf{b}_{1}\right)\right),\\
\label{eq3}
\end{equation}
where $\tau(\cdot)$ is the activation function (ReLU in our model), $W_{1}$ is the weight matrix, and $b_{1}$ is the bias vector at the first dense layer. Batch normalization ($\textup{BN}$) is used to train our model more efficiently. $h_{1}$ represents the output of the first layer. It is connected to the second layer using
\begin{equation}
\setlength\abovedisplayskip{2pt}
\setlength\belowdisplayskip{2pt}
\mu_{g} = \tau\left(\textup{BN}\left(\mathbf{W}_{2}^{T}h_{1} + \mathbf{b}_{2}\right)\right).\\
\label{eq4}
\end{equation}
Latent variables $\textup{z}_{g} \sim \mathcal{N}\left(\mu_{g},\sigma_{g}^{2}\right)$.
$\mu_{g}$ is the computed mean value of this Gaussian distribution. Similarly, $\sigma_{g}$ is computed by another 2-layer neural network with the same architecture as $\mu_{g}$. The latent vector $z_{g}$ is randomly sampled from $\mathcal N (\mu_{g},\sigma_{g})$.
The decoder architecture is also a 2-layer fully connected neural network. The decoded gene expression data is written as G':
\begin{equation}
\setlength\abovedisplayskip{2pt}
\setlength\belowdisplayskip{2pt}
G^{'} = \sigma \left(\textup{BN}\left(\mathbf{W}_{4}^{T}\left(\tau\left(\textup{BN}(\mathbf{W}_{3}^{T}z_{g} + \mathbf{b}_{3}\right)) + \mathbf{b}_{4}\right)\right)\right),
\label{eq5}
\end{equation}
where $\sigma$ represents sigmoid activation.
In our model, both the encoder and the decoder are 2-layer fully connected neural networks, with the architecture shown in Figure~\ref{fig GeneVAE}. The sizes of both encoder layers are set as 256, while the sizes of both decoder layers are set to match input data. When encoding gene expression data into latent vectors, we take $\mu_{g}$ as encoded features instead of sampling these vectors from a Gaussian distribution.

\subsection{JTVAE}
\label{appendix:JTVAE}
JTVAE consists of a graph VAE and a tree VAE. Molecules are decomposed as junction trees where nodes are valid molecular substructures. The decomposed junction tree is encoded with a tree VAE while the original molecular graph is encoded with a graph VAE. When generating molecules, the decoder of the tree VAE reconstructs the junction tree of the molecule, and the decoder of the graph VAE provides complementary connectivity information to reproduce the full molecular graph.

\subsubsection{Graph encoder}
The encoder of the graph VAE, which takes atoms as nodes in the graph, is implemented with a message passing network.\cite{gilmer2017neural} Messages pass from node to node for \textit{t} iterations. The final representation of each node is computed by aggregating its relevant messages from the message passing network, with these representations used to produce the final graph representation $\mathbf{h}_{G}$. The graph latent vector $\mathbf{z}_{G}$ is sampled from $\mathcal N(\mu_{G}, \sigma_{G})$, where $\mu_{G}$ and $\sigma_{G}$ are computed by 2 separate affine layers from the graph representation.

\subsubsection{Tree encoder}
The encoder of the tree VAE, in contrast, uses valid substructures of the molecule graph as nodes, and implements a message passing network based on a gated recurrent unit (GRU).\cite{chung2014empirical} The message $\mathbf{m}_{ij}$ passed from node $i$ to $j$ is updated as
\begin{equation}
\setlength\abovedisplayskip{2pt}
\setlength\belowdisplayskip{2pt}
\mathbf{m}_{ij} = \mathbf{GRU}(\mathbf{x}_{i}, \{\mathbf{m}_{ki}\}_{k \in N(i) \backslash j}),
\label{eq6}
\end{equation}
where $\mathbf{x}_{i}$ represents the type of substructure $i$, and $N(i)$ is the neighbor of $i$. Messages are passed from leaves to a randomly selected root and then from the root to leaves. After message passing, the tree representation $\mathbf{h}_{T}$ is produced by aggregating messages relevant to the root node. The tree latent vector $\mathbf{z}_{T}$ is sampled in a similar way with $\mathbf{z}_{G}$.

\subsubsection{Reconstruct molecules from latent vectors}
Using the given latent vectors $z_{G}$ and $z_{T}$, the tree VAE decoder generates a junction tree from $z_{T}$ first, and then the graph VAE decoder combines the substructures into the junction tree to produce the final reconstructed molecule.

The tree decoder starts from the root and traverses the junction tree in depth-first order recursively. It predicts the probability of the current node having children. Every time a child node is generated, the label of the child node is predicted. Nodes in the junction tree are labeled with the most likely valid substructure. The graph VAE decoder follows the order when the junction tree is reconstructed and only assembles one node at a time. While there may be multiple ways to assemble the substructures, JTVAE uses the highest scoring strategy.\cite{jin2018junction}
We make use of a JTVAE model pre-trained with the ZINC dataset. Similar to geneVAE, we use the predicted mean value of latent vectors as encoded features instead of sampling these vectors from a Gaussian distribution.

\subsection{Drug response prediction network}
\begin{figure}[!tpb]
\centering
\includegraphics[width=15cm]{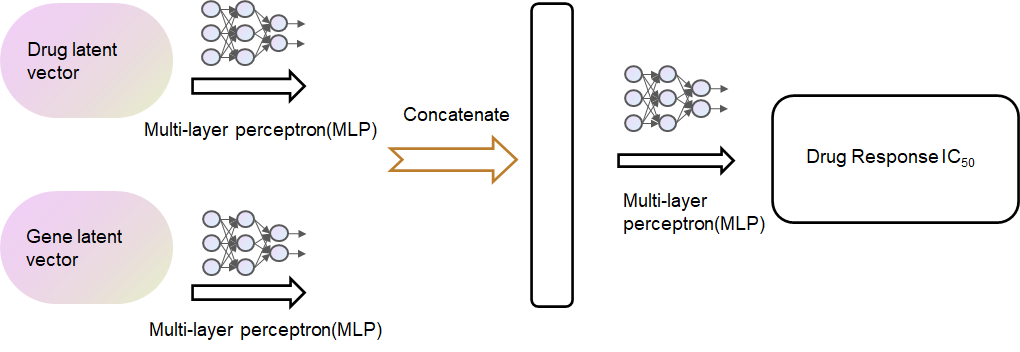}
\caption{The architecture of the drug response network to produce a final prediction. Two 3-layer MLP models post-process the encoded gene latent vectors and drug latent vector, and then another 4-layer MLP concatenates the output and produces a predicted $ln(IC_{50})$ value}
\label{fig MLP}
\end{figure}

As illustrated in Figure~\ref{fig MLP}, we implement two MLP models to post-process the latent features encoded by the two VAE models. We implement another MLP model to concatenate the processed output and produce the final drug response prediction. The input to the final MLP model is $\mathbf{a}_{all} =[\mathbf{a}_{gene}, \mathbf{a}_{drug}]$, where $\mathbf{a}_{gene}$ and $\mathbf{a}_{drug}$ are the outputs of the two post-processing MLP models. If $\mathbf{a}_{gene} \in \mathbb R^{d_{1}}$ and $\mathbf{a}_{drug} \in \mathbb R^{d_{2}}$, then $\mathbf{a}_{all} \in \mathbb R^{d_{1}+d_{2}}$, where $d_{1}$ is the dimensionality of $\mathbf{a}_{gene}$, and $d_{2}$ is the dimensionality of $\mathbf{a}_{drug}$. We compute the values of the perceptrons in the $i^{th}$ layer in the final MLP model according to
\begin{equation}
\setlength\abovedisplayskip{2pt}
\setlength\belowdisplayskip{2pt}
a_{all}^{i+1} = f^{'}(\mathbf{W}^{(i+1)T}a_{all}^{i} + \mathbf{b}^{i+1}),
\label{eq7}
\end{equation}
where $W^{(i+1)}$ is the weight matrix of the $i$-th layer in the final MLP model, and $f^{'}$ is a non-linear activation function. For the latter, we use the parametric rectified linear unit (PReLU) in our model. We complete the predicted $ln(IC_{50})$ in the last layer of the final MLP model according to
\begin{equation}
\setlength\abovedisplayskip{2pt}
\setlength\belowdisplayskip{2pt}
ln(IC_{50}) = f^{'}(\mathbf{W}^{(n)T}a_{all}^{n-1} + \mathbf{b}^{n}),
\label{eq8}
\end{equation}
where $n$ is the number of layers in the final MLP model.

In our model, both of the post-processing MLPs consist of 3-layer fully connected neural networks. Since the geneVAE and JTVAE are 256-dimension and 56-dimension vectors, respectively, we set the sizes of the two post-processing MLPs as (256, 256, 64) and (128, 128, 64). The final combining MLP is a 4-layer fully connected neural network with 128, 128, and 64 units in its hidden layers.

\subsection{Baseline model}
We substitute a support vector regression (SVR) network for MLP in our baseline model, showing a convenient way of using machine learning methods to make drug response predictions. We choose a poly kernel in our SVR model and set the parameter $C$ as 10.

\section{Experiments}
\subsection{Experiment set-up}
For our experiments, we trained geneVAE and JTVAE without supervision at the first stage. We used the pre-trained geneVAE to encode gene expression data filtered by the CGC data set or not on the breast cancer cell lines. We used JTVAE to encode anti-cancer drug molecular data. With these encoded features, we trained the baseline SVR model and MLP model for drug response prediction. We first tested our model on breast cancer cell lines followed by pan-cancer cell lines. Besides drug response prediction, we also demonstrated that our model generated effective drugs for given cancer cell lines. We split the training, validation and test sets in a 9:1:1 ratio for the SVR models, and the training, validation, and test sets in an 18:1:1 ratio for the MLP models. We implemented and debugged our models using PyCharm running under Microsoft Windows. We trained the model using an Nvidia GeForce RTX 2070 Super GPU.

\subsection{Pre-training geneVAE}
We aimed to minimize the sum of reconstruction and KL losses when training the geneVAE model. The reconstruction loss is $\mathcal L(G,G^{'})$, where $G$ represents initial input gene expression data and $G^{'}$ represents reconstructed data. The loss function could be either the mean squared loss {\ttfamily[MSEloss]} or the cross entropy loss {\ttfamily[CrossEntropyloss]}. We chose the cross entropy loss as the reconstruction loss in our experiments, since we normalized the input data and used sigmoid activation in the last layer to ensure that inputs and outputs were values between $0$ and $1$. Normalizing data is a very important aspect of gene expression studies and many commonly used normalization methods don’t treat samples independently but rather use the information contained in the whole dataset to guide the normalization procedure. This may lead to serious information leakage if the dataset is normalized before the training-test split. We normalize each row of the data, which is equivalent to normalize cancer cell lines of each gene. Therefore, we don't have to consider the effects of train, valid and test separation.

We pre-trained the geneVAE model on cancer cell line gene expression data both filtered by CGC dataset and without filtered. During training, we employed a warm-up strategy. The total VAE loss was set to
\begin{equation}
\setlength\abovedisplayskip{2pt}
\setlength\belowdisplayskip{2pt}
\mathbf{VAE\_Loss} = \mathcal L(G,G^{'})+\beta KL,
\label{eq9}
\end{equation}
where $KL$ is the KL loss and $\beta$ is a parameter that gradually increases from $0$ to $1$ during training. Because batchnorm layers were incorporated in geneVAE, we set the learning rate as $0.1$ initially for a faster learning. We also adopted a learning rate decay strategy in the training process, where the learning rate was multiplied by $0.8$ when validation loss fluctuated in a range of $0.5$ for over 10 epochs. The minimum learning rate was set as 0.01.

\begin{figure}[!tpb]
\centering
\begin{minipage}[t]{0.46\linewidth}
\centering
\includegraphics[width=6cm]{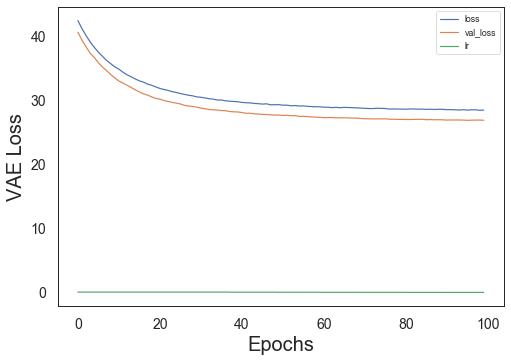}
\caption{VAEloss and lr with CGC}
\label{VAEloss+cgc}
\end{minipage}
\begin{minipage}[t]{0.46\linewidth}
\centering
\includegraphics[width=6cm]{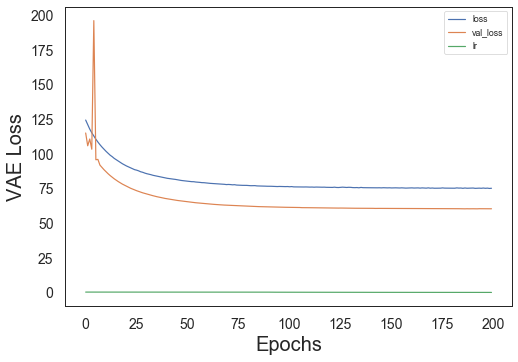}
\caption{VAEloss and lr without CGC}
\label{VAEloss+nocgc}
\end{minipage}
\end{figure}

In our tests, the total VAEloss (-ELBO) began to converge after approximately 100 epochs. As shown in Figures~\ref{VAEloss+cgc} and \ref{VAEloss+nocgc}, our model on CGC-selected gene expression data had an average VAEloss of 27.3, and the model without CGC selected gene expression data had an average VAEloss of 68 after the validation loss became stable.

\subsection{Exploring latent vectors from geneVAE}
\begin{figure*}[!htb]
\centering
\begin{minipage}[t]{0.8\textwidth}
\centering
\includegraphics[width=7cm]{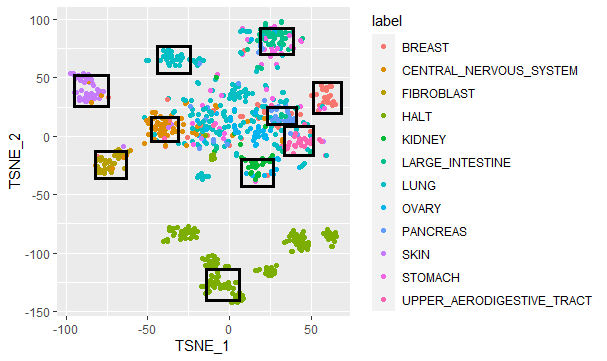}
\caption{T-SNE results of original gene expression data}
\label{tsne before geneVAE}
\end{minipage}
\begin{minipage}[t]{0.8\textwidth}
\centering
\includegraphics[width=7cm]{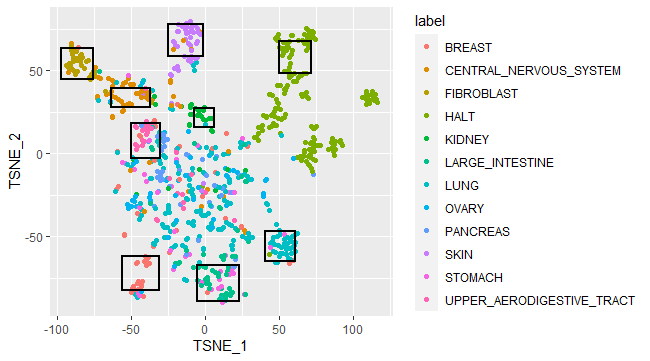}
\caption{T-SNE results of latent vectors encoded by geneVAE}
\label{tsne after geneVAE}
\end{minipage}
\end{figure*}

In this section, we demonstrate that the latent vectors encoded by geneVAE retained critical features of pan cancer gene expression data. We adopted the t-SNE method to reduce the dimensionality of both the original gene expression data (filtered by CGC dataset) and the latent vectors of gene expression data encoded by geneVAE, and we visualized them to reveal their similarity. We began by labelling the tissue type of each cell line, e.g., ``CERVIX'' or ``OVARY.'' We renamed "HAEMATOPOIETIC\_AND\_LYMPHOID\_TISSUE" as ``HALT for brevity. The parameters consisted of perplexity and the number of iterations for the single t-SNE model. We set perplexity to $n/120$, where $n$ is the number of cell lines, and the number of iterations to 3000. We further eliminated cancer types where the number of cancer tissues was below 30 for a better visualization result. Twelve main cancer types remained: {\ttfamily[BREAST, CENTRAL\_NERVOUS\_SYSTEM, FIBROBLAST, HALT, KIDNEY, SKIN, STOMACH]}. After removing tissues of rare cancer types, we visualize the data as in Figures~\ref{tsne before geneVAE} and \ref{tsne after geneVAE}. The results of the encoded latent vectors and those of the original data remained similar, where primary cancer tissue types (marked with black boxes) are separated clearly. Therefore, latent vectors encoded by geneVAE model retained the essential features of the original data. With geneVAE, our models were able to focus on the low-dimensional critical features of the original data and produce more accurate predictions.

\subsection{Exploring latent vectors from JTVAE}

\begin{figure}[!htpb]
\centering
\includegraphics[scale=0.5]{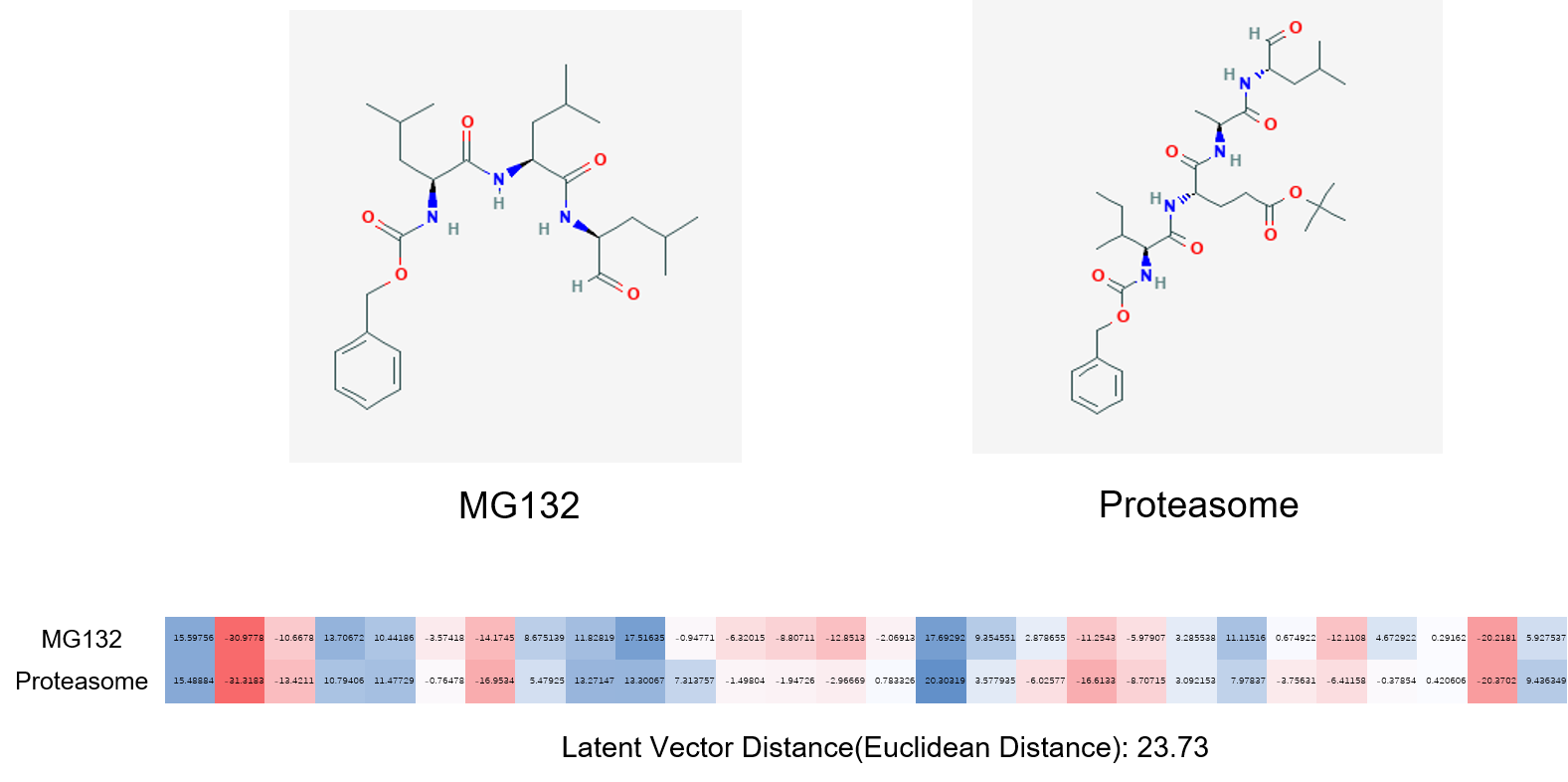}
\caption{MG132 and Proteasome, which are close in terms of the Euclidean distance between their latent vectors, share a majority of common functional groups}
\label{similar drugs}
\end{figure}

Many drugs having similar latent vectors encoded by JTVAE are also similar in their molecular structures. We measured the similarity of latent vectors of different drugs in terms of Euclidean distance. Shorter distances indicate a higher similarity between two drug latent vectors. For example, MG132 (inhibitor) and Proteasome (inhibitor) share a short Euclidean distance between their latent vectors of about 23.73. We obtained their molecular structures from the Pubchem database and found that they share a majority of functional groups, as shown in Figure~\ref{similar drugs}. Small differences were found in a carboxyl group and an amide at the ends of the molecules.

Although many drugs are similar in their latent vectors, their performance varies when used against different cancer cell lines. However, our drug prediction network captured these subtle differences and produced accurate predictions. We focused on the example of MG132 and Proteasome used against the HCC1187 cancer cell line. We removed these two pieces of data from the training set, and tested our trained model on them. The predicted $ln(IC_{50})$ of MG132 and Proteasome in cell line HCC1187 were 0.84 and $-0.866$ in our best model, while the actual $ln(IC_{50})$ values of these two drugs are 1.589 and $-0.181$, respectively. Although the predicted values were not very close to the expected ones, our models did not confuse these two samples. Therefore, despite strong drug similarities, our drug prediction network still differentiated each of them and produced reasonable results.

\section{Future work}
Since the usefulness of filtering out a gene subset with the CGC dataset is demonstrated by our experiments, we think it worth exploring the importance of selecting representative gene subsets. More promising methods such as network propagation based on the STRING protein-protein interaction database could be used to improve our model further.\cite{oskooei2018paccmann} We also wish to explore incorporating attention mechanism-based models to improve performance.\cite{manica2019toward} Recent work using graph neural networks (GNN) shows the potential of GNNs in dealing with drug molecular data. A combination of VAE and GNN (VGAE)\cite{kipf2016variational} could be adopted for this problem. VGAE would take advantage of VAE model to be generative and incorporate the GNN to process graph data efficiently. We believe a modified version of the originally proposed VGAE\cite{kipf2016variational} is a promising way to predict drug response and generate new drugs. Finally, since our model performs well on drug response prediction with good potential for drug discovery, we would like to build a toolkit based our model.

\end{document}